\documentclass[letterpaper, 10 pt, conference]{ieeeconf}  
\IEEEoverridecommandlockouts       

\usepackage{graphics} 
\usepackage{epsfig} 
\usepackage{mathptmx} 
\usepackage{times} 
\usepackage{amsmath} 
\usepackage{amssymb}  
\usepackage{subfigure}
\usepackage{algorithm} 
\usepackage{algpseudocode}
\usepackage[colorlinks=false]{hyperref}
\usepackage{multirow} 
 
\begin{document}

\title{\LARGE \bf
Learning NEAT Emergent Behaviors in Robot Swarms}

\author{Pranav Rajbhandari$^{1}$ and Donald Sofge$^{2}$
\thanks{$^{1}$Pranav Rajbhandari is the corresponding author and with Department
of Computer Science, Carnegie Mellon University, Pittsburgh, PA, USA, {\tt\small prajbhan@alumni.cmu.edu}. }%
\thanks{$^{2}$Donald Sofge is with the Naval Research Laboratory, Washington D.C., USA.}%
}

\maketitle

\newcommand{\figref}[1]{Fig.~\ref{#1}}
\newcommand{\tabref}[1]{Table~\ref{#1}}
\newcommand{\algoref}[1]{Algorithm~\ref{#1}}
\newcommand{\todo}[1]{\textcolor{red}{\\TODO: {#1}\\}}
\newcommand{\lrquote}[1]{\lq#1\rq}
\newcommand{\commentout}[1]{}

\def\fullfiguresize{.92\linewidth}
\def\halffiguresize{.49\linewidth}

\begin{abstract}
When researching robot swarms, many studies observe complex group behavior emerging from the individual agents' simple local actions. 
However, the task of learning an individual policy to produce a desired group behavior remains a challenging problem.
We present a method of training distributed robotic swarm algorithms to produce emergent behavior. Inspired by the biological evolution of emergent behavior in animals, we use an evolutionary algorithm to train a \lrquote{population} of individual behaviors to produce a desired group behavior. 
We perform experiments using simulations of the Georgia Tech Miniature Autonomous Blimps (GT-MABs) aerial robotics platforms conducted in the CoppeliaSim simulator. 
Additionally, we test on simulations of Anki Vector robots to display our algorithm's effectiveness on various modes of actuation.
We evaluate our algorithm on various tasks where a somewhat complex group behavior is required for success. These tasks include an Area Coverage task and a Wall Climb task. 
We compare behaviors evolved using our algorithm against \textit{designed policies}, which we create in order to exhibit the emergent behaviors we desire.
\end{abstract}


\section{Introduction}
\label{intro}
\subsection{Emergent Behavior}
Emergent behavior is a phenomenon observed in swarms of agents. It is generally defined as a complex swarm behavior which occurs as a consequence of each individual agent following a relatively simple control scheme \cite{emergent_def_1, emergent_def_2}.
Examples of this can be found in the behavior of groups of animals, such as how some species of fire ants have been observed to create rafts out of their bodies to survive flooding \cite{ant_raft}. 
Emergent behavior is also exhibited in migrating swarms of some species of caterpillars, which walk on top of each other in order to create a \lrquote{rolling swarm} \cite{caterpillar_swarm}. This allows the caterpillars to migrate quicker than simply walking. 
Similar instances of complex emergent behavior have been observed in the field of swarm robotics.  

\subsection{Swarm Robotics}
\label{swarm_robotics}
Though there are various definitions of swarm robotics, it is generally agreed that robot swarms are multi-agent systems where each agent is autonomous and has low complexity \cite{securing_emergence_comm, security_swarm_robotics, swarm_robot_intro}. They are used for various tasks, including exploration and surveillance. They are characterized by using locally communicating distributed systems as opposed to a central controller. Due to this, they remain functional as the swarm size is increased. However, they do not have a central controller, so careful fine tuning of the individual policies is required to achieve a desired swarm behavior. 

It is well known that emergent behavior can arise in robot swarms, and past studies have explored this. Pagello et al. \cite{cooperative_task_emergence} study how to make a robot swarm perform a cooperative task by creating emergent behaviors. They implement this by dynamically assigning predefined roles to each robot. A function $Q$ is learned for each agent, which takes in local information and outputs the best role for the agent to adopt. After refining their $Q$ functions, they observe cooperative emergent behavior in their chosen task, robot soccer.

In a more recent study, Oliveri et al. \cite{continuous_learning_emergence} use a Monte Carlo scheme to continuously refine the behavior of individual agents in a swarm. They tested their method on robots connected in a line, which were each able to push away from their neighbors. Their study concluded that training each agent to optimize its individual velocity resulted in the entire group of connected robots crawls forward.



\subsection{Evolutionary Algorithms/NEAT}
Evolutionary algorithms are 
algorithms inspired by biological evolution, where a user defined \textit{fitness function} is optimized by repeatedly transforming a \lrquote{population} of solutions by spawning new members similar to members with the best fitness.
We inspect NeuroEvolution (NE), a type of evolutionary algorithm that evolves neural networks. In early NE algorithms, the topology\footnote{The topology of a neural network refers to the graph structure of the nodes and the connections between them} of the neural network is fixed, and the mutations take place in the weights of the connections \cite{neuroevolution}. However, using the Neuroevolution of Augmenting Topologies (NEAT) algorithm, the topolocy of the neural network can evolve as well, theoretically allowing generalization to problems of arbitrary complexity \cite{NEAT}.

\subsection{Organization}

The rest of the paper is organized as follows: Section \ref{related_work} discusses related work. Section \ref{methodology} details our proposed learning algorithm, as well as common sensing and output schemes we use. We present our experiments in Section \ref{experiments}, and discuss the results in Section \ref{results}. The conclusions we draw are in Section \ref{conclusion}.

\section{Related Work}
\label{related_work}
Much research has focused on the goal of learning individual behaviors for a robot swarm \cite{
learning_complex_swarm_behavior, 
securing_emergence_comm,
swarm_learning_comparision,
automode,
communication_for_deep_swarm_learning,
continuous_learning_emergence,
cooperative_task_emergence}.
Behjat et al. \cite{learning_complex_swarm_behavior} explore how to learn tactical swarm behavior through a combination of various techniques. These techniques include learning neural network based robot policies, dynamically organizing the swarm into groups, and Pareto filtering of points of interest to reduce the problem dimensionality.

Fan et al. \cite{swarm_learning_comparision} compare different swarm intelligence algorithms against each other to evaluate their relative performance in obstacle performance under different circumstances. 
The algorithms they evaluate are the bat algorithm (BA), particle swarm optimization (PSO), and the grey wolf optimizer (GWO). They find that PSO outperforms BA, which outperforms GWO in general. However, GWO performs better than the other two algorithms in the case of large swarms and large communication ranges.

The AutoMoDe algorithm, designed by Francesca et al., defines an agent policy as a finite state machine whose states are preexisting \textit{constituent behaviors} \cite{automode}. The optimization process applies the F-race algorithm \cite{frace} to learn transitions between constituent behaviors based on features observed from the environment. In their initial paper, they successfully train robots to perform an aggregation task.

Dorigo et al. \cite{communication_for_deep_swarm_learning} explore how to create communication protocols between agents in a swarm that best help methods such as deep reinforcement learning create good decentralized control policies.

\thispagestyle{empty}
\pagestyle{empty}

In addition to the studies by Oliveri and Pagello which focus on the learning aspect (described in Section \ref{swarm_robotics}), there has also been research on helping the stability of emergent behaviors in swarms. In 2022, Chen and Ng explore secure communications between robots in a swarm \cite{securing_emergence_comm}. They model these communications as a series of random graphs, and use a method involving hash chains to identify \lrquote{rogue robots} with high probability. This allows them to ensure that the emergent behavior exhibited by the swarm is protected. 
In their paper, they also create a system to distinguish different classes of robot swarms by identifying differences in the robot homogeneity, the interactions between robots, and the interactions with a central control.

Our work is closely related to research done by Trianni et al. in 2003 \cite{evolving_aggregate_behaviors}. They use an evolutionary algorithm to evolve an \lrquote{aggregation} behavior in a swarm of robots, inspired by the self-organized aggregation behavior observed in the cellular slime mold \textit{Dictyostelium discoideum} \cite{slime_slug}. Trianni's evolutionary algorithm uses a neural network with no hidden layers, and evolves by mutating the weights. They use a fitness function that seeks to minimize the mean distance of each agent from the swarm's center of mass. 
Later research by Bahceci explores this further by varying the parameters of the evolutionary algorithm to see how it affects the evolution of aggregation behaviors \cite{evolving_aggregation_behaviors_case_study}.

We expand on Trianni's idea by generalizing the algorithm to allow user-defined fitness functions. Additionally, we use the NEAT algorithm as opposed to simply evolving the weights of a set network. These extensions allow our algorithm to theoretically evolve a network to arbitrary complexity in order to create a desired emergent behavior. 

We use the GT-MABs as the robots for our experiments \cite{GTMAB}. The control system for their actuation was designed based on the dynamic model created in \cite{blimp_dynamics, blimp_autopilot}.
\section{Methodology}
\label{methodology}
\subsection{Evolutionary Algorithm}
We propose NEAT as a candidate for learning emergent behavior in robot swarms. We will evolve a population of neural networks, evaluated through the performance of a homogeneous\footnote{Each agent is controlled with a copy of the same neural network} robot swarm in one episode.

Our algorithm requires as input a user-defined fitness function. This function acts on a full episode of a robotic swarm's behavior, and returns a real number evaluating the swarm behavior.

The proposed algorithm does the following in a loop:
\begin{itemize}
    \item 
    For each network $x$ in a population of neural networks, a robot swarm is initialized such that each member contains a copy of $x$ for control.
    \item A full episode is run, and the output of the fitness function is used as the fitness of $x$.
    \item After fitnesses are collected, the next generation of networks is evaluated with NEAT.
\end{itemize}
We implement this on model robot swarms in a CoppeliaSim simulator \cite{COPPELIA}. For learning, we use the NEAT-Python package \cite{neat_python}.
We use ROS2 Foxy to handle transferring sensing and actuation messages between CoppeliaSim and Python \cite{ROS_FOXY}. Our full implementation is available on Github \cite{swarm_coppeliasim}.

\subsubsection{Drawbacks/Justification}
The main drawback of this method is each full episode generates one fitness score for evaluation. This is quite wasteful in comparison to alternative methods like Reinforcement Learning (RL), which use each timestep as a training example.

However, using RL would require a user-defined evaluation function to assign a value to each action of each agent based on its benefit to the swarm behavior. This seems like quite a strong restriction, and the difficulty of defining this function would greatly limit the applications of RL to this problem. 

Thus, the sample inefficiency of using an evolutionary algorithm can be justified by how applicable the algorithm is.

\subsection{Network Inputs}
The primary method for control of robot swarms is to have each agent act independently on local information. This method allows generalization of learned policies to varied swarm sizes. Thus, we generally use local observations of the environment and of nearby agents for inputs to each agent's policy. 

Since the position of other agents in the swarm is usually vital to training a good policy, we specify two sensing schemes that we use for our experiments.

\begin{figure}[htbp]
    \centering
\subfigure[Distance Sense]{
\includegraphics[width=\halffiguresize]{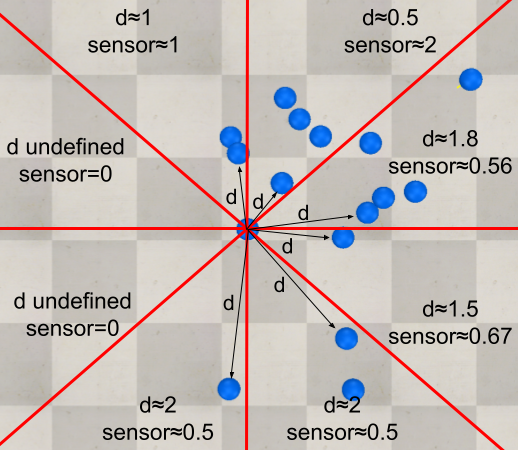}
}\subfigure[Neighbor Sense]{
\includegraphics[width=\halffiguresize]{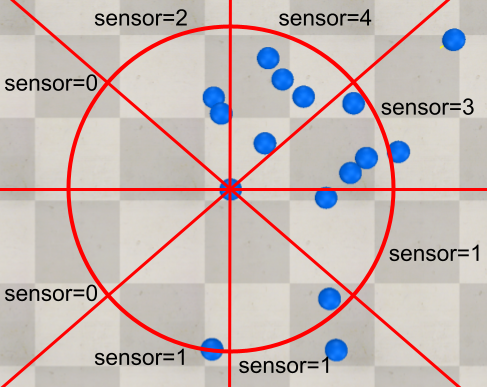}
}
\caption{$k$-tant sensing scheme with $k=8$}
\label{input_scheme}
\end{figure}

We define \textit{$k$-tant} sensing, where we split the environment into $k$ regions based on direction relative to the agent that is sensing. We allow the agent to either sense the distance to the nearest neighbor\footnote{We invert the distance so that an empty region corresponds to a sensor value of 0} or the number of neighbors in each region, as displayed in \figref{input_scheme}. Although we only use 2-D experiments, this sensing scheme very easily generalizes to 3-D (and $N$-D) sensing by discretizing the relative angles to each neighboring agent.

\subsection{Network Outputs/Agent Actuation}

Since the NEAT algorithm outputs vectors of any specified dimension, this algorithm can very easily be fed into a velocity controller, or even into a low-level motor controller. In our experiments, we use both methods to verify the robustness of our algorithm to output scheme.

\section{Experiments}
\label{experiments}

\begin{figure}[H]
    \centering
\subfigure[Anki Vector]{
\includegraphics[width=\halffiguresize]{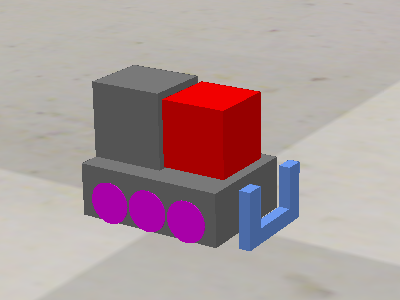}
}\subfigure[GT-MAB]{
\includegraphics[width=\halffiguresize]{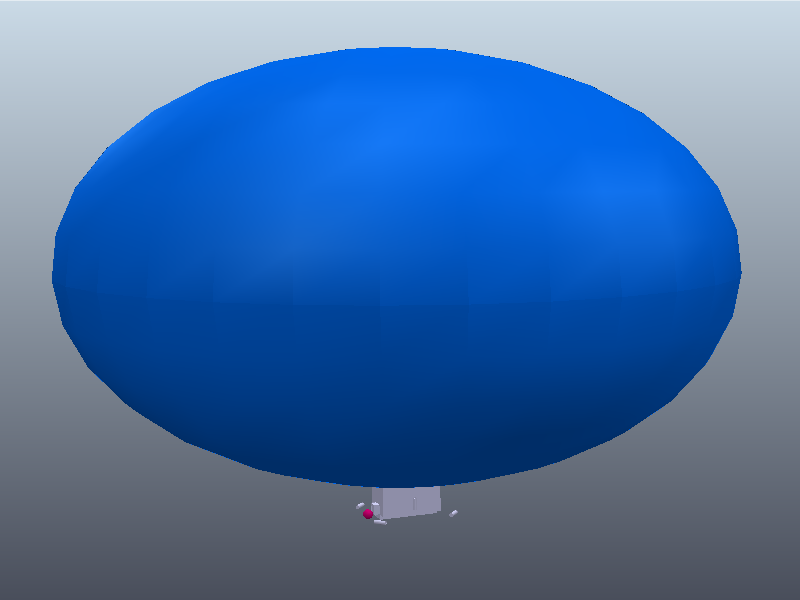}
}
\caption{CoppeliaSim models}
\label{robot_images}
\end{figure}

To test our algorithm, we use swarms of GT-MAB robots \cite{GTMAB} and Anki Vector robots \cite{ANKI}, simulated in CoppeliaSim (\figref{robot_images}). We feed the neural network outputs into a velocity controller for the GT-MABs and directly into wheel velocity controllers for the Anki robots. We define various tasks for the robot swarms to complete, and train our evolutionary algorithm for 50 generations in CoppeliaSim.
We terminate each episode after 60 seconds, and compare the best evolved policy's performance against a \textit{designed policy} created to solve each task.

\subsection{Task: Area Coverage}
\begin{figure}[htbp]
\centerline{\includegraphics[width=\fullfiguresize]{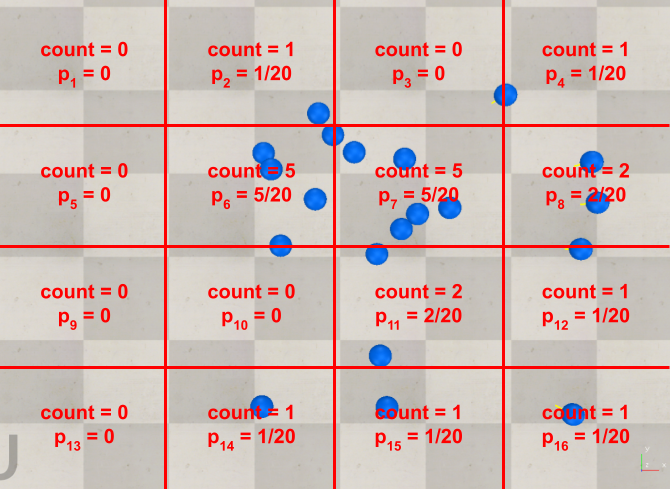}}
\caption{Calculation of deployment entropy}
\label{deploy_entropy}
\end{figure}

In this task, we simulate a \lrquote{search and rescue} scenario. The environment we use is a square arena with the agents spawned randomly near the center (displayed in \figref{area_prog}).
For the fitness function, we adapt deployment entropy, a measure of how well distributed the agents become in the environment \cite{deployment_entropy}. Deployment entropy is defined by discretizing the environment into a grid, and measuring the entropy of the distribution of agents in that grid. In the example of \figref{deploy_entropy}, we would calculate this value as $\sum\limits_i-p_i\log(p_i)\approx 2.05$.
Since entropy is maximized by a uniform distribution, we believe using this as a fitness function would encourage the agents to spread out in the environment, the desired behavior for a \lrquote{search and rescue} mission.  We implement this task for the both the Anki robots and the GT-MABs.

For the GT-MABs, we use 20 agents and divide the environment into a grid with 16 units for calculating deployment entropy. We chose $\lfloor 20\rfloor^2=16$ as this makes 1-2 agents per square optimal. For input, we use $k$-tant Distance Sensing with $k=8$. We also allow the walls to be sensed on each $k$-tant, since this lets the agents avoid crashing into them.

For the Anki Vectors, we use 10 agents (as this number would comfortably fit in our environment) and divide the environment into $\lfloor\sqrt{10}\rfloor^2=9$ units for deployment entropy. For input, we use $k$-tant Distance Sensing ($k=8$) along with of the Anki's onboard proximity sensor.

For both agents, we define a \textit{designed policy} where each agent moves away from the closest neighbors. We expect this to cause the agents to distribute themselves evenly, producing the desired swarm behavior.

\subsection{Task: Wall Climb}

In previous research we established that a swarm of GT-MABs was able to climb a wall that was much taller than the altitude they were assigned to hold \cite{wall_climbing}. Upon closer inspection, we realized that that collisions were resulting in the agents registering each other as the floor, effectively stacking their desired heights. A stack of three agents would allow the top one to pass over the wall.

We also note that the \lrquote{wall climbing} behavior is susceptible to the sensing angle of the GT-MAB's ultrasound-based range sensor. With a large sensing angle, the GT-MABs are able to sense the wall itself when they are next to it, which results in a single agent being able to climb the wall on its own. To avoid this, we use a narrow ultrasound angle so that the agents must stack in order to climb the wall. 

\begin{figure}[htbp]
    \centering
\subfigure[Evolved behavior]{
\includegraphics[width=\halffiguresize]{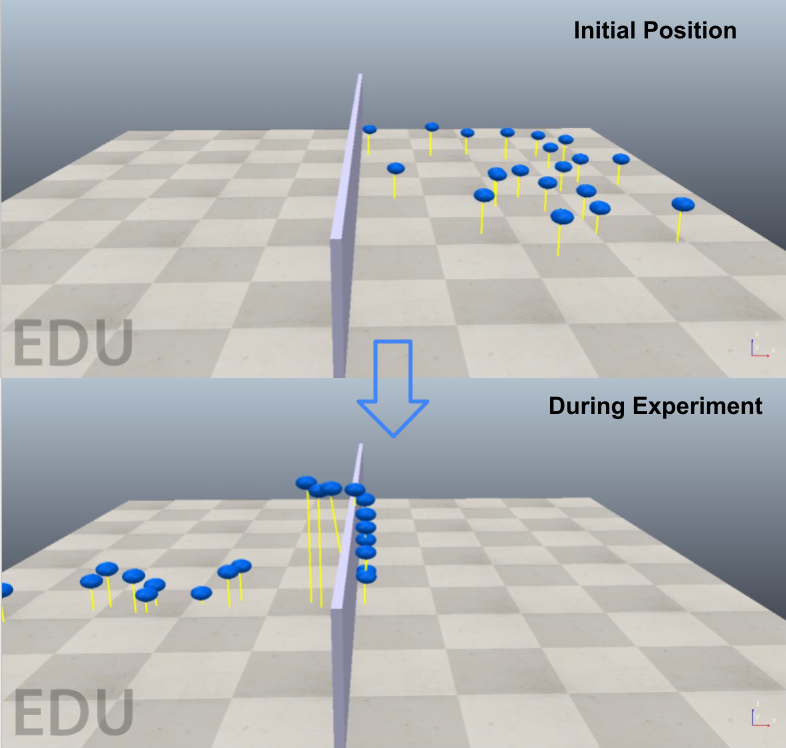}
}\subfigure[\textit{Designed policy} behavior]{
\includegraphics[width=\halffiguresize]{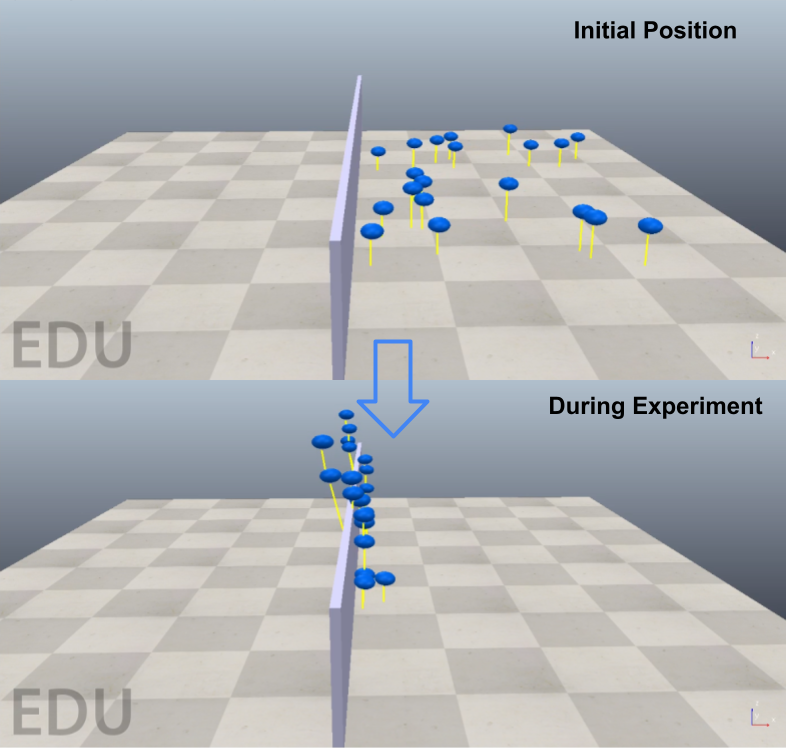}
}
\caption{Neighbor Sensing Wall Climb behaviors}
\label{wall_climb_prog}
\end{figure}

We design the Wall Climb task to replicate this scenario.
The environment we use is an arena with a 3 meter tall wall on the $y$ axis. A swarm of 20 GT-MABs spawn randomly on the right side, as shown in \figref{wall_climb_prog}. 
For the fitness function, we use the number of agents that end the experiment on the other side of the wall. If no agents succeed, we subtract a penalty of the distance of the closest blimp to the wall, to speed up initial exploration. We believe using this as a fitness function will encourage the emergent \lrquote{aggregation} behavior, since this is the only way the agents can climb the wall.

For our input scheme, we experiment with both $k$-tant Distance Sensing and $k$-tant Neighbor Sensing, with $k=8$. We compare the two experiments to display robustness of our algorithm to input scheme.

We define a \textit{designed policy} where each agent moves towards either their closest neighbor or the direction with the most neighbors (depending on the input scheme), as well as slightly towards the direction of the wall. We expect this to cause the GT-MABs to flock and stack on top of one another, producing the desired swarm behavior of flocking over the wall. For our \textit{designed policy}, we expect that Neighbor Sensing will perform slightly better, as this will encourage the agents to seek the largest group of neighbors.

\commentout{
\subsection{Task: Surround Target}

In this task, we simulate a goal of securing a perimeter around a target \cite{perimeter_defense}. 
The environment we use is an arena containing a cylinder target spawned randomly, with 6 GT-MABs also spawned randomly in the arena, as in \figref{target_surround_prog}.

For the fitness function, we consider all the agents close to the target. Using polar coordinates centered at the target, we find the measure of the largest uncovered arc $a$, and set the fitness to $-a$. We also subtract a small penalty of distance to target to encourage initial exploration. We believe using this as a fitness function would encourage the agents to \lrquote{secure the perimeter} since minimizing the largest uncovered arc would minimize the furthest distance any point on the perimeter is from some agent.

We define a \textit{designed policy} where each agent moves towards the target, as well as slightly away from their closest neighbor. We expect this to cause the GT-MABs to head for the target, then slowly spread out, producing the desired swarm behavior.
}
\commentout{
\subsection{Task: Maze Solving}
\begin{figure}[htbp]
\centerline{\includegraphics[width=\fullfiguresize]{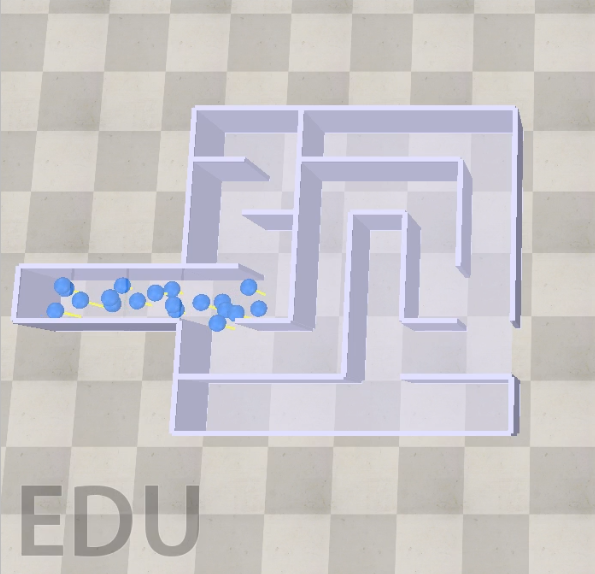}}
\caption{Maze Solving experiment initial position}
\label{maze_solve_init}
\end{figure}
In this task, we generate a random maze with the goal of having the swarm solve it. The environment includes a cell to hold the blimps next to the starting square. The GT-MABS spawn randomly in the cell and starting square, as shown in \figref{maze_solve_init}. To generate the maze, we use the randomized Recursive Backtracking algorithm, implemented with Pymaze \cite{pymaze}. We force the entry and exit squares to be on opposite ends. To create the fitness function, we find the maze distance $m$ of the closest agent to the exit square of the maze, then use $-m$ as the fitness since we want to minimize this.
We use 20 agents and terminate each experiment after 60 seconds, but also allow the experiment to terminate early with a perfect fitness of 0 if any agents reach the exit square.

We include Octant Distance Sensing as our input (\figref{input_scheme}), with the caveat of not being able to see neighbors through walls. We also include a 4-vector encoding the walls of the cell the agent is in. 
We use our GT-MAB actuation scheme as output (\figref{output_schemes}).

We define a \textit{designed policy} where each agent moves away from their closest visible neighbors, with the exception of never being able to move towards a wall of its cell. We expect this to cause the agents to distribute themselves evenly and fill the maze, producing the desired swarm behavior.
}

\section{Results}
\label{results}
\def\num{4}

\begin{table}[htbp]
\caption{Fitness in Evolved$^{\mathrm{\boldsymbol{a}}}$ and {Designed Behaviors}}
\begin{center}
\begin{tabular}{|c|c|c|c|}
\hline
\textbf{Experiment}&\textbf{Behavior}& \textbf{Mean}& \textbf{Stdev.}
\\
\hline
\hline
\multirow{2}{*}{\textbf{\textit{GT-MAB Area Coverage}}}& Evolved & 2.53 & 0.10 
\\ 
\cline{2-\num} 
\textbf{\textit{}} & Designed & 2.62 & 0.12 
\\
\hline
\hline

\multirow{2}{*}{\textbf{\textit{Anki Area Coverage}}}& Evolved & 2.09 & 0.083 
\\
\cline{2-\num} 
\textbf{\textit{}} & Designed & 1.87 & 0.20 
\\
\hline
\hline

\textbf{\textit{Wall Climb}}& Evolved & 16.70 & 1.12 
\\
\cline{2-\num} 
\textbf{\textit{(Distance Sense)}} & Designed & 12.13 & 3.78 
\\
\hline
\hline

\textbf{\textit{Wall Climb}}& Evolved & 16.72  & 1.32 
\\
\cline{2-\num} 
\textbf{\textit{(Neighbor Sense)}} & Designed & 15.33 & 1.58  
\\
\hline
\hline

\multirow{2}{*}{\textbf{\textit{Surround Target}}}& Evolved & -2.21 & .70 
\\
\cline{2-\num} 
\textbf{\textit{}} & Designed & -1.24 & 0.14 
\\
\hline
\multicolumn{\num}{l}{$^{\mathrm{\boldsymbol{a}}}$: Using the best genome from the final generation.}
\end{tabular}
\label{comparision_table}
\end{center}
\end{table}

For each experiment, after training for 50 generations, we compare the fitness of the best genome in the final generation with the fitness of our \textit{designed policy}. We run each for 60 trials in their respective experiments and collect the fitness mean and standard deviation. We report our results in \tabref{comparision_table}.


\subsection{Area Coverage}

\begin{figure}[htbp]
\centerline{\includegraphics[width=\fullfiguresize]{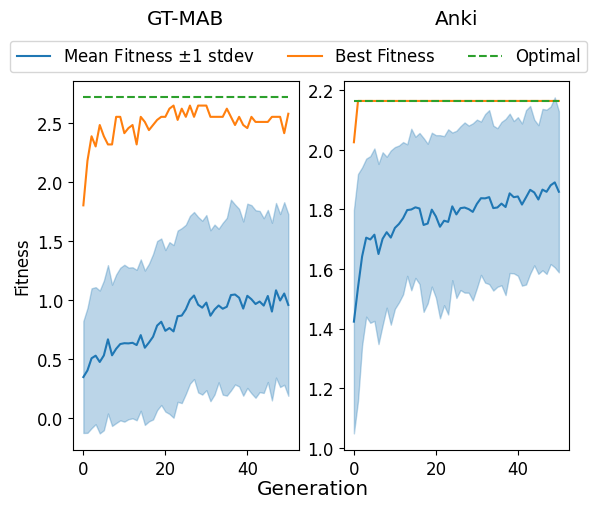}}
\caption{Area Coverage population fitness across generations}
\label{area_fitness}
\end{figure} 

\begin{figure}[htbp]
    \centering
\subfigure[GT-MAB NEAT]{
\includegraphics[width=\halffiguresize]{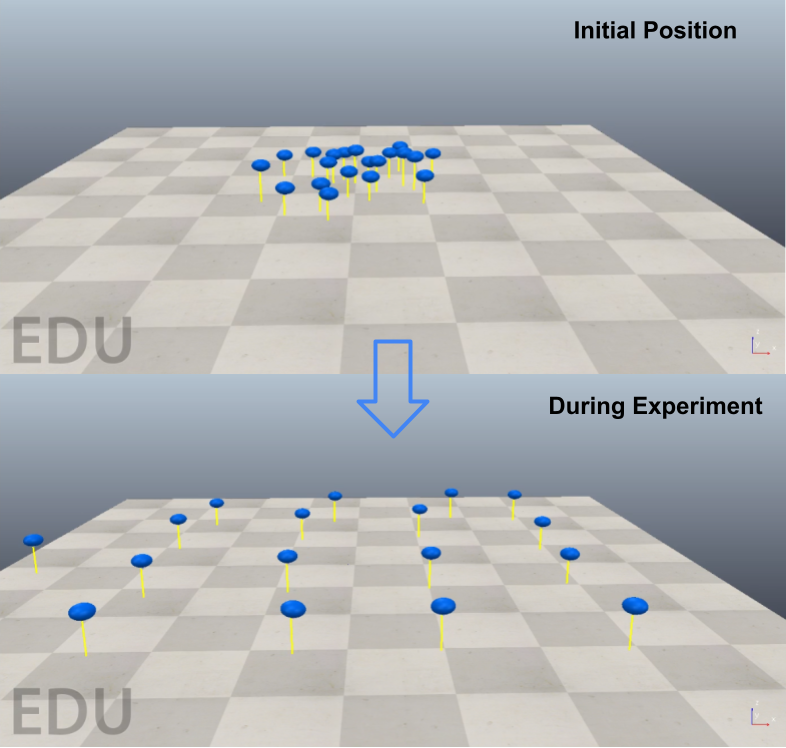}
}\subfigure[GT-MAB \textit{designed policy}]{
\includegraphics[width=\halffiguresize]{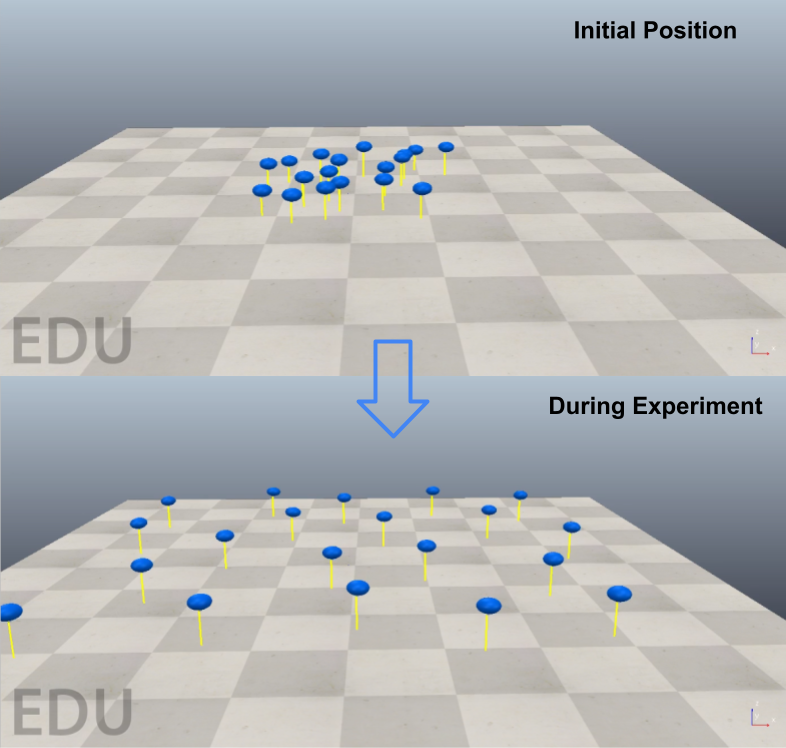}
}
\subfigure[Anki NEAT]{
\includegraphics[width=\halffiguresize]{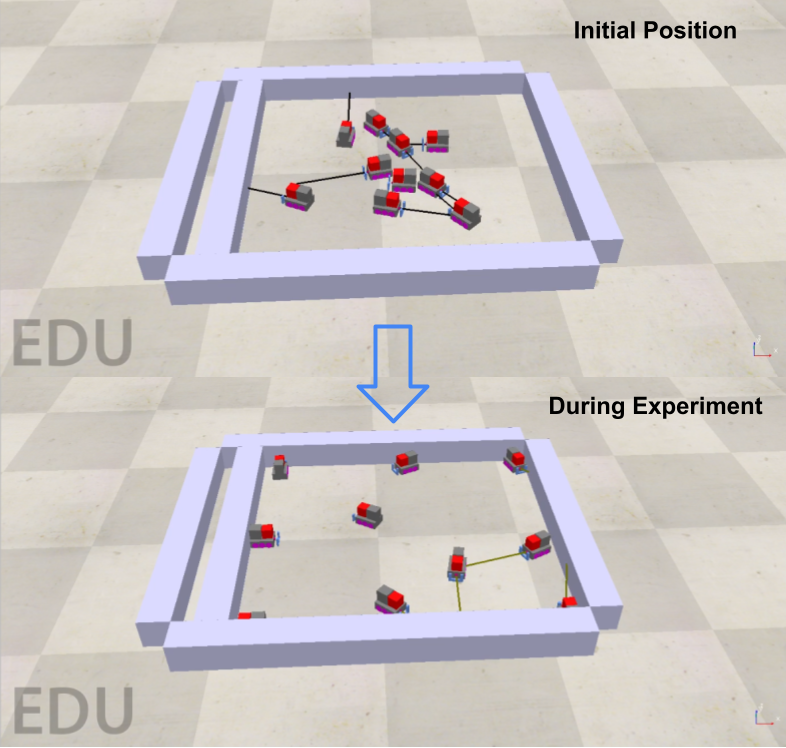}
}\subfigure[Anki \textit{designed policy}]{
\includegraphics[width=\halffiguresize]{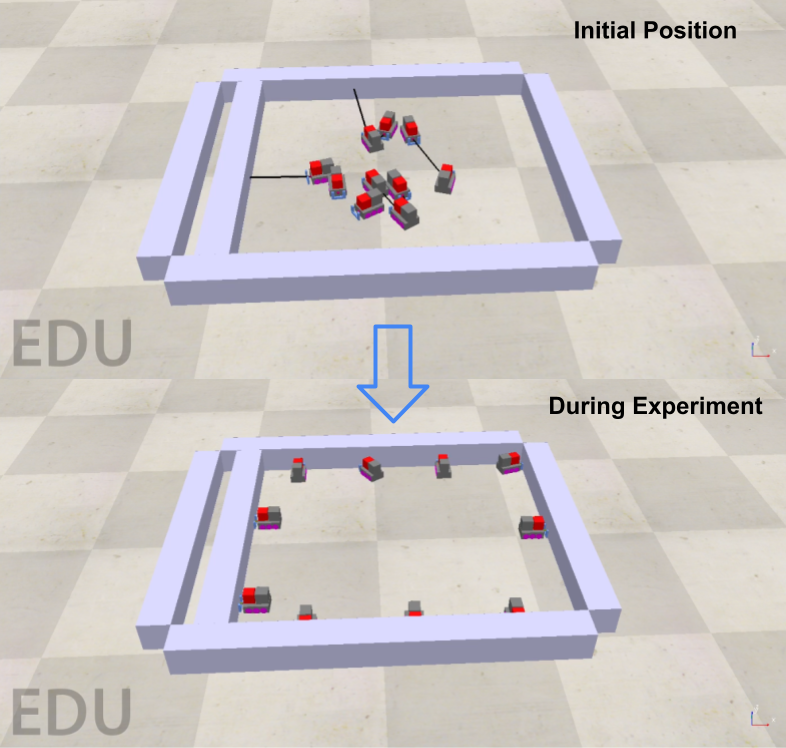}
}
\caption{Area Coverage behaviors}
\label{area_prog}
\end{figure}

After training in CoppeliaSim, we observe that the evolved behavior for both GT-MABs and Ankis seems to be for each agent to move in a direction away from its close neighbors. The GT-MABs accomplish this directly through the velocity controller, while the Ankis learned to spin in circles and reverse whenever their proximity sensor sensed an agent in front of them. For both experiments, the result seems to be a well distributed swarm (\figref{area_prog}). From \figref{area_fitness}, we can see that both experiments generated agents that achieved close to the theoretic maximal entropy\footnote{2.718 for GT-MABs from placing one agent in 12 of the squares, and two agents in 4 of the squares; 2.16 for Ankis by placing one agent in 8 of the squares and two agents in 1 square}, with the Ankis achieving it almost immediately.  

The result of the \textit{designed policy} (\figref{area_prog}(b, d)) also had the swarm distribute in the environment. 
Comparing these results in \tabref{comparision_table}, we can see that for GT-MABs, the \textit{designed policy} outperforms the evolved behavior by about one standard deviation. For the Anki experiment, the opposite is true, with the evolved behavior outperforming the \textit{designed policy} by about one standard deviation. 
This seems to be due to the \textit{designed policy} favoring sending agents to the edges of the environment.

Overall, we show that in this task, our algorithm learns a local behavior that closely approximated the desired \lrquote{search and rescue} emergent swarm behavior. The evolved behavior greatly outperforms the \textit{designed policy}, which shows that our evolutionary algorithm performs comparably to designing the behavior with knowledge of the task.

\subsection{Wall Climb}

\begin{figure}[htbp]
\centerline{\includegraphics[width=\fullfiguresize]{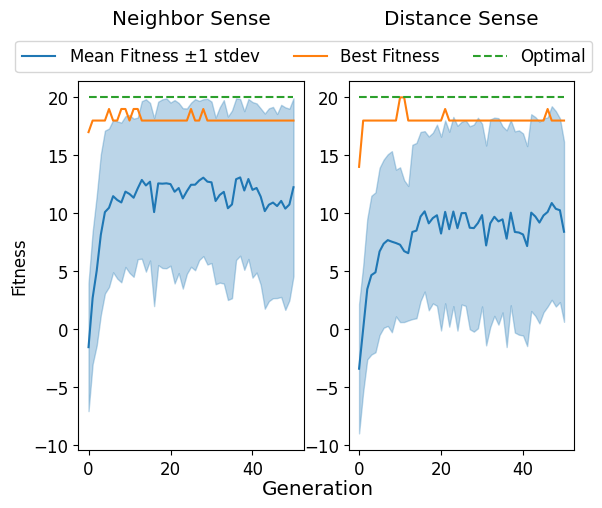}}
\caption{Wall Climb population fitness across generations}
\label{both_wall_climb_fitness}
\end{figure} 

\begin{figure}[htbp]
    \centering
\subfigure[Evolved behavior]{
\includegraphics[width=\halffiguresize]{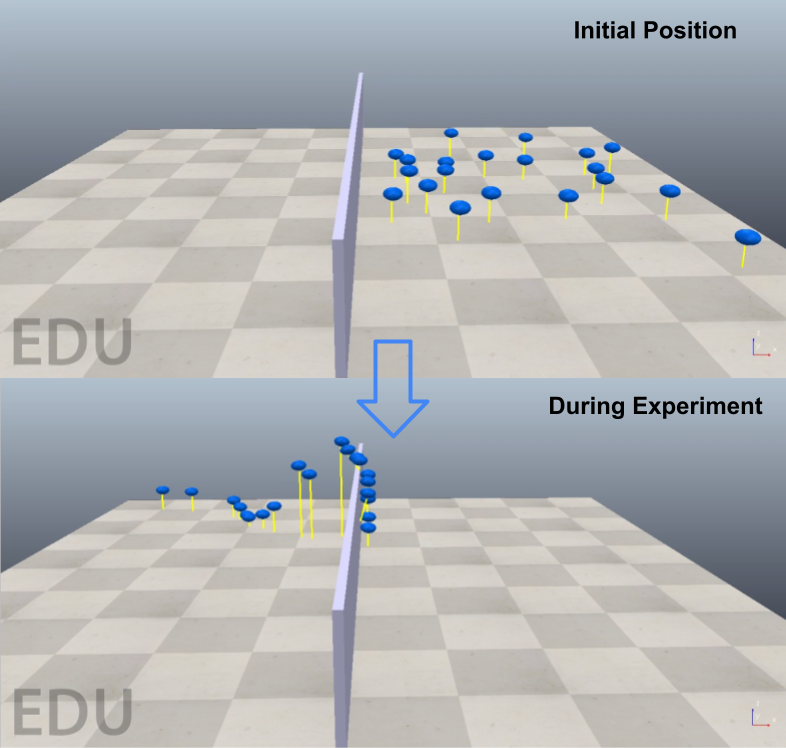}
}\subfigure[\textit{Designed policy} behavior]{
\includegraphics[width=\halffiguresize]{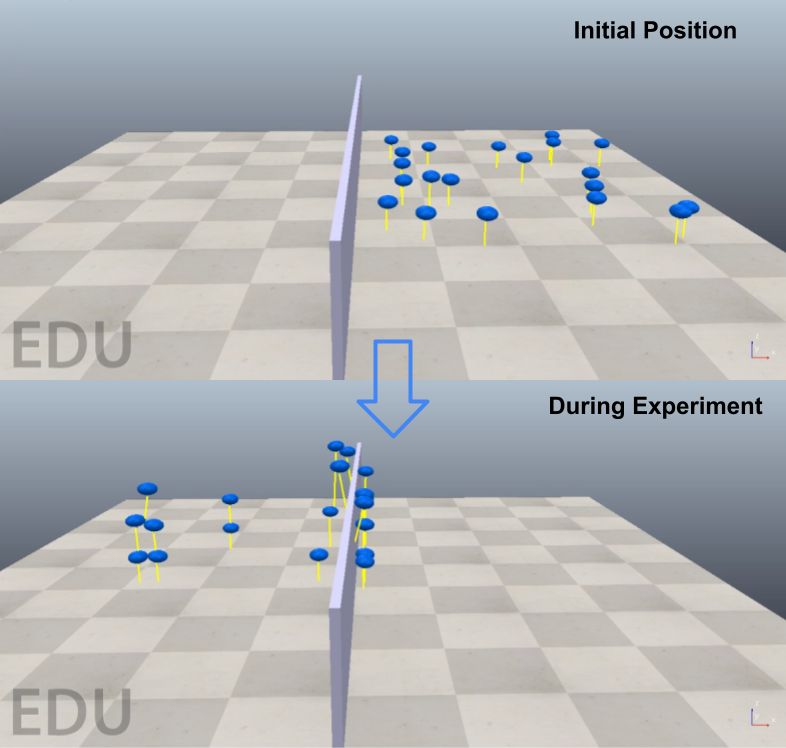}
}
\caption{Distance Sensing Wall Climb behaviors}
\label{dist_wall_climb_prog}
\end{figure}

After training the GT-MAB models in CoppeliaSim, we observe that the evolved behavior for both experiments seems to be for each agent to move in a direction towards its closest neighbors, in addition to moving in the direction to climb the wall.
The result of this behavior does seem to be a flock of agents stacked on top of each other, climbing the wall. 
From \figref{both_wall_climb_fitness}, we can see that after a few generations, the best genome of each generation achieved having about 18 GT-MABs make it over the wall. 

We noticed that both sensing methods achieved similar population fitness values across generations \figref{both_wall_climb_fitness}. From \tabref{comparision_table}, we see there is no statistically significant difference between the two fitness results.\footnote{We perform a two sample $t$-test with unequal variance on the fitnesses obtained from the evolved Neighbor Sensing and Distance Sensing experiments. Using their respective means and standard deviations of (16.70,1.12) and (16.72,1.32) with a sample size of 60 for both, we arrive at a $p$-value of $p>.9$, which is much larger than $.05$, the accepted value for statistical significance. This shows that with our sample size, there is no significant difference between the fitnesses obtained from the different sensing methods.}

The result of the \textit{designed policy} in both of these experiments similarly causes the agents to flock together and climb the wall. However, in this case we do notice a difference in the two modes of sensing. In the Distance Sensing experiments, the \textit{designed policy} seems to be more likely to cause the agents to form several smaller groups as opposed to one large one. This results in a lower fitness due to more agents being left behind, as shown in \tabref{comparision_table}. 
Comparing this with our evolved behavior, we can see in both experiments, the evolved behavior outperforms the best \textit{designed policy} by at least one standard deviation. 

Overall, we show that in this task, our algorithm learns a local behavior that closely approximated the desired \lrquote{flocking} emergent swarm behavior. We can further see that although the different sensing modes has an effect on our \textit{designed policy}, the evolutionary algorithm is robust to these variations.

\commentout{
\subsection{Surrounding a Target}

\begin{figure}[H]
    \centering
\subfigure[Evolved behavior]{
\includegraphics[width=\halffiguresize]{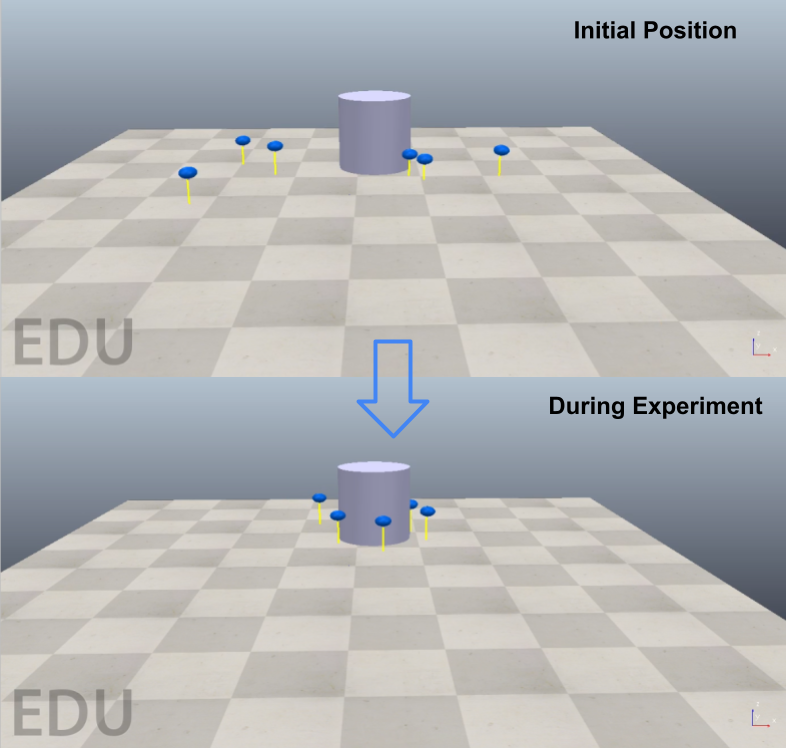}
}\subfigure[\textit{Designed policy} behavior]{
\includegraphics[width=\halffiguresize]{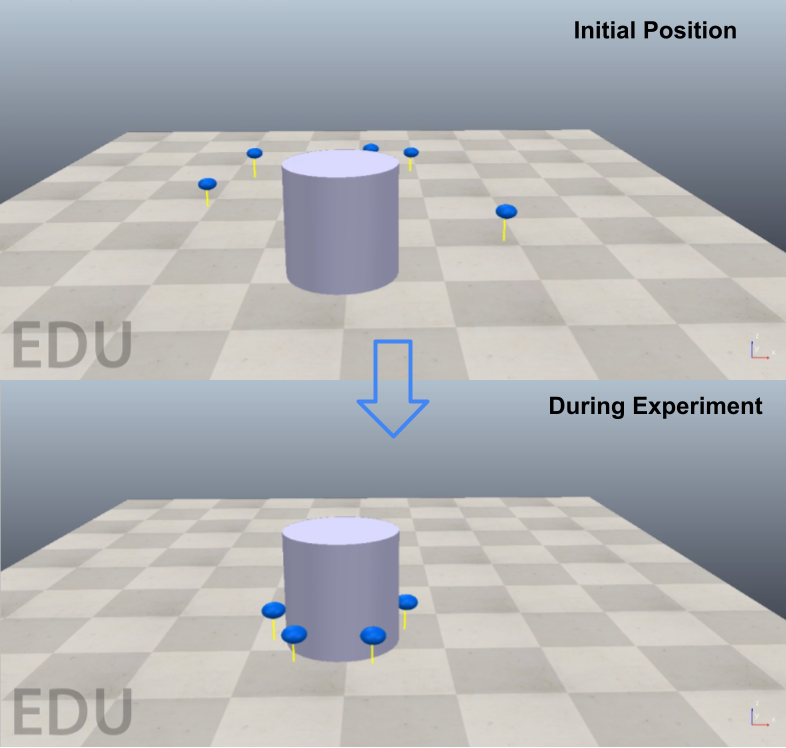}
}
\caption{Surround Target behaviors}
\label{target_surround_prog}
\end{figure}

\begin{figure}[htbp]
\centerline{
\includegraphics[width=\fullfiguresize]{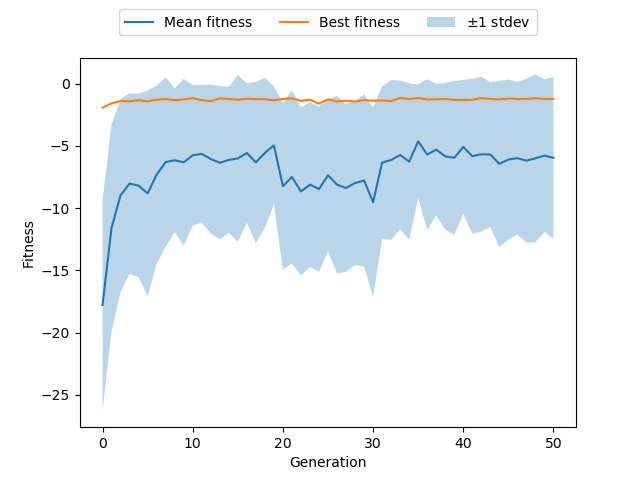}

}
\caption{Surround Target population fitness across generations}
\label{target_surround_fitness}
\end{figure}

\commentout{
\begin{figure}[htbp]
\centerline{
\includegraphics[width=\halffiguresize]{evolution_plots/6_blimp_1_cylinder_obstacle_surround_range_4_0_all.png}
\includegraphics[width=\halffiguresize]{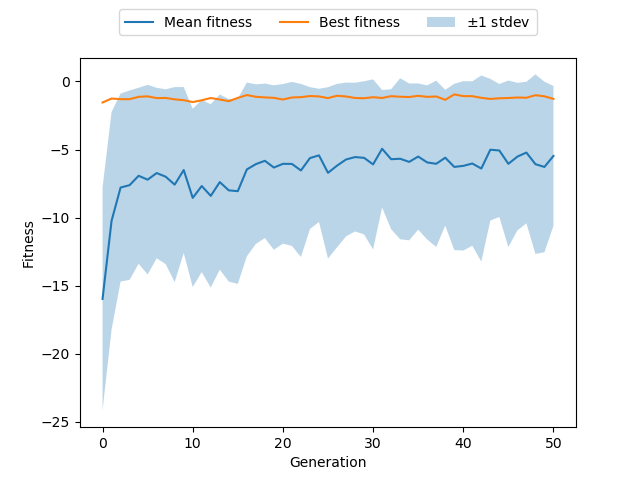}
}
\caption{Surround Target population fitness across generations (cylinder target on left, cube target on right)}
\label{target_surround_fitness}
\end{figure}
}

After training the GT-MAB models in CoppeliaSim, we observe that the behavior they evolved seems to be for each agent to spiral around the target, in addition to moving away from its closest neighbor.
\figref{target_surround_prog} shows an experiment run with the network with the best genome in the last generation of training.
The result of this behavior does seem to be a flock of agents surrounding the target. 
From \figref{target_surround_fitness}, we can see that after a few generations, the best genome of each generation achieves having a largest uncovered arc of about 2 radians.  

The result of the \textit{designed policy}, shown in \figref{target_surround_prog}, also caused the agents to surround the target. Comparing these results, we can see that our \textit{designed policy} outperforms the evolved behavior by at least one standard deviation (\tabref{comparision_table}). 

Overall, we show that in this task, our algorithm learns a local behavior that closely approximates the desired emergent swarm behavior. However, our \textit{designed policy} still does outperform the evolved behavior, which suggests that for some tasks it is more difficult to evolve the optimal behavior. We believe fine tuning of the evolution parameters may help this issue, but that is outside the scope of our research.

}
\section{Conclusion}
\label{conclusion}
In this paper, we present a novel extension of the NEAT algorithm designed to learn emergent behaviors in robot swarms. The algorithm we present is robust to robot swarms with various modes of sensing and actuation. 
Results from simulations show that individual agent behaviors evolved using this method are comparable to hand designed policies at producing desired complex emergent behaviors. 

In future research, we plan to test our evolved policies on the physical GT-MABs and Anki Vector robots. We also plan to evaluate our algorithm on a more complex set of tasks. We also may explore the fine tuning of NEAT parameters to try improving our results. 
Additionally, the accuracy of the CoppeliaSim simulator seems to be dependant on the performance of the computer, as this determines the speed that each agent can respond to stimuli. 
In our experiments, we choose to speed up our experiments by running multiple simulations in parallel, which potentially caused delays in agent response time. In future experiments, less parallelization, a more powerful computer, or a different simulator could improve the stability of our algorithm.

\bibliographystyle{plain} 
\bibliography{refs}

\end{document}